\PassOptionsToPackage{unicode}{hyperref}
\PassOptionsToPackage{hyphens}{url}
\documentclass[
]{article}
\usepackage{amsmath,amssymb}
\usepackage{lmodern}
\usepackage{ifxetex,ifluatex}
\ifnum 0\ifxetex 1\fi\ifluatex 1\fi=0 
  \usepackage[T1]{fontenc}
  \usepackage[utf8]{inputenc}
  \usepackage{textcomp} 
\else 
  \usepackage{unicode-math}
  \defaultfontfeatures{Scale=MatchLowercase}
  \defaultfontfeatures[\rmfamily]{Ligatures=TeX,Scale=1}
\fi
\IfFileExists{upquote.sty}{\usepackage{upquote}}{}
\IfFileExists{microtype.sty}{
  \usepackage[]{microtype}
  \UseMicrotypeSet[protrusion]{basicmath} 
}{}
\makeatletter
\@ifundefined{KOMAClassName}{
  \IfFileExists{parskip.sty}{%
    \usepackage{parskip}
  }{
    \setlength{\parindent}{0pt}
    \setlength{\parskip}{6pt plus 2pt minus 1pt}}
}{
  \KOMAoptions{parskip=half}}
\makeatother
\usepackage{xcolor}
\IfFileExists{xurl.sty}{\usepackage{xurl}}{} 
\IfFileExists{bookmark.sty}{\usepackage{bookmark}}{\usepackage{hyperref}}
\hypersetup{
  pdftitle={Textwash - automated open-source text anonymisation},
  pdfauthor={Bennett Kleinberg; Toby Davies; Maximilian Mozes},
  hidelinks,
  pdfcreator={LaTeX via pandoc}}
\usepackage[margin=1in]{geometry}
\usepackage{longtable,booktabs,array}
\usepackage{calc} 
\usepackage{etoolbox}
\makeatletter
\patchcmd\longtable{\par}{\if@noskipsec\mbox{}\fi\par}{}{}
\makeatother
\IfFileExists{footnotehyper.sty}{\usepackage{footnotehyper}}{\usepackage{footnote}}
\makesavenoteenv{longtable}
\usepackage{graphicx}
\makeatletter
\def\maxwidth{\ifdim\Gin@nat@width>\linewidth\linewidth\else\Gin@nat@width\fi}
\def\maxheight{\ifdim\Gin@nat@height>\textheight\textheight\else\Gin@nat@height\fi}
\makeatother
\setkeys{Gin}{width=\maxwidth,height=\maxheight,keepaspectratio}
\makeatletter
\def\fps@figure{htbp}
\makeatother
\providecommand{\tightlist}{%
  \setlength{\itemsep}{0pt}\setlength{\parskip}{0pt}}
\ifluatex
  \usepackage{selnolig}  
\fi
\newlength{\cslhangindent}
\newlength{\csllabelwidth}
\newenvironment{CSLReferences}[2] 
 {
  \setlength{\parindent}{0pt}
  \ifodd #1 \everypar{\setlength{\hangindent}{\cslhangindent}}\ignorespaces\fi
  \ifnum #2 > 0
  \setlength{\parskip}{#2\baselineskip}
  \fi
 }%
 {}
\usepackage{calc}

\title{Textwash - automated open-source text anonymisation}
\author{Bennett Kleinberg\footnote{\href{mailto:bennett.kleinberg@tilburguniversity.edu}{\nolinkurl{bennett.kleinberg@tilburguniversity.edu}}} \and Toby
Davies \and Maximilian Mozes}
\date{Version: 27 August, 2022}

\begin{document}
\maketitle
\begin{abstract}
The increased use of text data in social science research has benefited
from easy-to-access data (e.g., Twitter). That trend comes at the cost
of research requiring sensitive but hard-to-share data (e.g., interview
data, police reports, electronic health records). We introduce a
solution to that stalemate with the open-source text anonymisation
software \emph{Textwash}. This paper presents the empirical evaluation
of the tool using the TILD criteria: a technical evaluation (how
accurate is the tool?), an information loss evaluation (how much
information is lost in the anonymisation process?) and a
de-anonymisation test (can humans identify individuals from anonymised
text data?). The findings suggest that Textwash performs similar to
state-of-the-art entity recognition models and introduces a negligible
information loss of 0.84\%. For the de-anonymisation test, we tasked
humans to identify individuals by name from a dataset of crowdsourced
person descriptions of very famous, semi-famous and non-existing
individuals. The de-anonymisation rate ranged from 1.01-2.01\% for the
realistic use cases of the tool. We replicated the findings in a second
study and concluded that Textwash succeeds in removing potentially
sensitive information that renders detailed person descriptions
practically anonymous.
\end{abstract}

\hypertarget{introduction}{%
\section{Introduction}\label{introduction}}

With the increasing digitisation of society and human communication,
text data are becoming more important for research in the social and
behavioural sciences (Gentzkow, Kelly, and Taddy 2019; Salganik 2019).
Advances made in natural language processing (NLP) in particular have
led to exciting insights derived from text data (e.g., on emotional
responses to the pandemic (Kleinberg, Vegt, and Mozes 2020) or on the
rhetoric around immigration in political speeches (Card et al. 2022);
for an overview, see (Boyd and Schwartz 2021)). Importantly, the use of
computational techniques to quantify and analyse text data has triggered
a demand, especially for large datasets (often of several tens of
thousands of documents) that can be harnessed for machine learning
approaches (e.g., (Socher et al. 2013; Lewis et al. 2020)). That status
quo of a need for larger datasets and an appetite to use text data for
the study of social science phenomena has resulted in a dilemma: many of
the important questions require targeted, primary data collection or
access to potentially sensitive data. However, such data are hard to
obtain, not because they do not exist but because sharing them is
constrained by data protection regulations and ethical concerns. One
potential consequence is that research activity may be biased toward
topics for which suitable data is more readily available rather than
those most important.

One of the few viable solutions to this dilemma is automated text
anonymisation; that is, the large-scale processing of text data so that
individuals cannot be identified from the resulting output. Such a
method would allow for the flow of sensitive data so that the staggering
potential of text data can be exploited for scientific progress. With
this paper and the tool it introduces, we seek to enable researchers to
work with such sensitive data in a way that protects the privacy of
individuals whilst retaining the usefulness of anonymised data for
computational text analysis.

\hypertarget{anonymising-text-data}{%
\subsection{Anonymising text data}\label{anonymising-text-data}}

Text anonymisation refers to redacting - and potentially replacing -
personally identifiable information (PII) within text data. Since such
information is the crucial concern of data protection regulations --
data is protected if and only if it can be associated with a living
individual -- its removal means that such data can be freely shared.
Text anonymisation aims to facilitate the sharing of data whilst
protecting the identities of individuals that are the subject of text
data. Text anonymisation comes with two critical challenges.

First, for subsequent content-based text analyses, an anonymisation tool
should anonymise text so that the anonymised text remains readable,
meaningful, and useful for syntax- and content-based analyses after
anonymisation (i.e.~it should preserve the semantic context of the
text). Consider, for example, the sentence ``Joe Biden is the current
president of the United States.'' If we replace all PII with a generic
term (e.g., XXX), we obtain the sentence ``XXX XXX is the current XXX of
the XXX XXX,'' rendering the sentence semantically meaningless and
context-free. Any subsequent text analysis would fail to capture the
context, semantic roles or other relevant linguistic constructs.

Second, evaluating the performance of text anonymisation tools is
challenging. Even if we assume an anonymisation tool to identify almost
all PII, just a few unidentified sensitive words can reveal an
individual's identity and jeopardise the whole anonymisation procedure.
For example, if the sequence above were anonymised as ``XXX XXX is the
current president of the United States,'' the reader would be able to
infer with an educated guess that the person mentioned in the text is
Joe Biden.

In an attempt to pave the way for privacy-preserving data sharing, in
this paper, we introduce \emph{Textwash}, an open-source text
anonymisation tool that anonymises text data in a fully automated way.
Textwash anonymises texts in a semantically-meaningful manner, ensuring
that the anonymised documents remain usable for downstream text analyses
with respect to a document's syntactic properties and content. The tool
achieves this in a two-stage process, consisting of i) the automatic
identification of relevant information from input documents and ii) the
subsequent replacement of the detected information with
meaning-preserving tokens. Rather than replacing PII with XXX, we use
category-specific replacement tokens for each identified word. For
example, Joe would be replaced with {[}firstname1{]}, Biden with
{[}lastname1{]}, president with {[}occupation1{]}, and United States
with {[}location1{]}, resulting in ``{[}firstname1{]} {[}lastname1{]} is
the current {[}occupation1{]} of the {[}location1{]}.''

Textwash is based on supervised machine learning techniques, leveraging
pre-trained contextualised word representations as provided by the BERT
language model (Devlin et al. 2018). In order to train a model capable
of automatically identifying relevant information in the input text, we
first annotate a large corpus of text data sourced from the British
National Corpus (BNC; Consortium and others (2007)) as well as the Enron
email dataset.\footnote{ \url{https://www.cs.cmu.edu/~enron/}} We then
fine-tune a pre-trained BERT model on the collected data.

\hypertarget{existing-approaches-to-text-anonymisation}{%
\subsection{Existing approaches to text
anonymisation}\label{existing-approaches-to-text-anonymisation}}

Machine learning-based text anonymisation approaches have been proposed
in various languages. Mamede, Baptista, and Dias (2016), for example,
propose an approach based on named entity recognition (NER) and
coreference resolution to anonymise texts in Portuguese. Their model
provides three modes of anonymisation: suppression (i.e., each PII is
replaced with a generic token such as XXX), tagging (i.e., each PII is
replaced with a category-specific and indexed token, such as
{[}\textbf{\emph{ORGANIZATION123}}{]}), and random substitution (i.e.,
each PII is replaced with a random PII of the same category). The
authors evaluate the performance of their model with respect to an
automated precision- and recall-based method as well as through human
evaluation. However, in contrast to Textwash, their proposed model only
focuses on the entities person, location, and organisation, and hence
might miss crucial PII corresponding to other categories, as we show in
this paper.

Elsewhere, NETANOS (named entity-based text anonymization for open
science) has been proposed as an anonymisation tool for the English
language (Kleinberg and Mozes 2017). NETANOS utilises an available named
entity recognition tool, the Stanford Named Entity Tagger (Finkel,
Grenager, and Manning 2005), to identify PII from input text data.
Unlike Textwash, NETANOS does not require annotated training data but,
as a consequence, only identifies persons, locations, organisations, and
dates.

More recently, Francopoulo and Schaub (2020) approached text
anonymisation from the context of customer relationship management
(CRM). The authors suggest a method comprising an NER-based module, an
entity linker, and a substitution module evaluated on a collection of
French legal and administrative documents. In contrast to Textwash, an
off-the-shelf NER module, Tagparser (Francopoulo 2007), is used.

Adams et al. (2019) proposed AnonyMate, which classifies identifiable
information as either PII or CII (corporate identifiable information)
and removes these. Their method is based on an analysis of historical
chat logs, from which 24 entity types of interest were extracted. Based
on these entities, the authors propose an annotated NER data set
comprising the six languages English, German, Swedish, Spanish, Italian,
and Swedish. They then trained two NER models, one based on conditional
random fields (CRF) and one based on recurrent neural networks, and
combined the NER model with a coreference resolution model. The proposed
approach is evaluated against automatic performance metrics (precision,
recall, F1-score), leaving the question of how well their method would
work when tested against a human benchmark.

While the tools above rely on identifying to-be-removed words, Hassan et
al. (2019) devised an anonymisation method based on word embeddings.
Their approach, however, is limited in that documents can only be
anonymised if a specific entity is to be removed, and hence does not
generalise to the task described in this paper. Various related methods
have been proposed recently (Di Cerbo and Trabelsi 2018; Berg,
Chomutare, and Dalianis 2019; Romanov et al. 2019).

The tool that comes closest to Textwash's aim is
\href{https://scrubadub.readthedocs.io/}{scrubadub} -- a Python package
using various existing off-the-shelf packages (e.g., spaCy, Stanford NER
detector) to detect PII in text, albeit with fewer categories than
Textwash and without retaining context between replaced entities.
Unfortunately, similar to AnonyMate, the tool is not evaluated against
de-anonymisation or information loss, leaving its usefulness for
data-sharing activities unclear.

In addition to these published examples, some commercial tools claim to
perform text anonymisation. However, these tools typically do not
provide empirical evidence of their performance and are closed-source.
This lack of transparency and the absence of evidence of their validity
essentially disqualifies them from scientific applications.

The present paper provides a novel perspective on text anonymisation
that addresses one or several weaknesses of previous approaches. Most
immediately, we evaluate Textwash according to rigorous empirical
criteria that surpass those used in previous work: as well as measuring
its technical performance, we evaluate its performance on the core task
of anonymisation via a realistic scenario involving human participants.
In addition, the software is developed and tested\footnote{Note that
  some commercial tools exist that fail to provide any information about
  the validity of their software and the process of building it and
  thereby disqualify from an application for scientific purposes.}
Specifically, with a focus on scientific research purposes: Textwash
adheres to the principles of open science (i.e., open source code,
transparent processes, free non-commercial use) and is usable without an
internet connection (i.e., not imposing any vulnerabilities when
processing sensitive data).

\hypertarget{aims}{%
\subsection{Aims}\label{aims}}

This paper has two aims: introducing the Textwash tool for automated
text anonymisation and providing an empirical evaluation of the tool. We
first detail the software and then report two validation studies that
put the tool to various tests.

\hypertarget{transparency-statement}{%
\subsection{Transparency statement}\label{transparency-statement}}

The Textwash software is an open-source Python project and is available
and documented at \url{https://github.com/maximilianmozes/textwash}. All
data collected for the reported studies and the material used are
publicly available in that repository. All participants in the studies
where data were collected provided informed consent, and the procedures
were approved by the IRB at University College London.

\hypertarget{the-textwash-tool}{%
\section{The Textwash tool}\label{the-textwash-tool}}

\hypertarget{requirements-for-modern-text-anonymisation}{%
\subsection{Requirements for modern text
anonymisation}\label{requirements-for-modern-text-anonymisation}}

For the development of Textwash, we identified a set of requirements
enumerated in this section.

\begin{itemize}
\tightlist
\item
  A text anonymisation tool operating on sensitive data should be usable
  offline, on a regular computer (e.g., laptop), without sending any
  data through external APIs for processing and anonymisation. This is
  to ensure that Textwash adheres to potential privacy regulations that
  users have in place and to remove any requirement to trust a third
  party.
\item
  Anonymised text data need to retain value for secondary analysis in
  NLP (e.g., topic modelling, sentiment analysis or coreference
  resolution) since one of the main goals of this software is to provide
  researchers and practitioners with a tool to share anonymised datasets
  for further data analysis.
\item
  A text anonymisation tool should be generalisable to new contexts and
  domains. While rule-based approaches might be capable of anonymising
  textual documents corresponding to a specific topic, Textwash aims to
  be as adaptable as possible and should be able to anonymise
  out-of-domain texts. In order to do this, Textwash is based on machine
  learning-based methods and uses linguistic patterns extracted from
  contextual information to predict whether individual words and phrases
  contain sensitive information.
\item
  For the software to be trustworthy and auditable, its mechanisms
  should be transparent, and for users to fully exhaust its
  capabilities, it should be customisable. The software should therefore
  be open source and available for public use.
\item
  A tool that deals with the sensitive context of anonymisation needs to
  be validated appropriately using empirical experiments. These should
  go beyond the automated metrics typically used and should involve
  testing by a set of human judges.
\end{itemize}

\hypertarget{potentially-sensitive-information}{%
\subsection{Potentially sensitive
information}\label{potentially-sensitive-information}}

The often-used definition of personally identifiable information rests
on the assumption that a few pre-defined categories capture all that
could lead to the identification of an individual. But, as this paper
will show, there are pieces of information that - in combination with
other information - could reveal an identity even if they would not
count as personally identifiable information in themselves. These
include references to hyper-specific attributes of a person (e.g., a
mention of a specific pronunciation error that someone makes) and
context-dependent sensitive information (e.g., mentioning that an actor
performed in the last film of a famous series). Put differently: the
information that can reveal an individual's identity is not limited to
categories such as dates, names and locations. Therefore, we propose a
concept encompassing the full spectrum of information that could reveal
an identity: \textbf{potentially sensitive information (PSI)}. PSI is
hereafter used to describe any piece of information that could directly
or indirectly (e.g., through the combination with other information) be
useful to identify an individual.

With that broadening of identification risk in mind, the task for a text
anonymisation system can be summarised as identifying PSI in an input
sequence and replacing that information with generic terms that do not
reveal any information about individuals mentioned in the text.

\hypertarget{identifying-potentially-sensitive-information}{%
\subsection{Identifying potentially sensitive
information}\label{identifying-potentially-sensitive-information}}

The first problem of identifying PSI input words can be described as a
named entity recognition (NER) task using supervised learning
techniques. Given a corpus of text sequences where PSI words are
annotated as such, we can train a machine learning model to classify
tokens in input sequences according to whether they represent PSI or
not.

Textwash realises this by utilising pre-trained contextualised word
representations using BERT. Specifically, we fine-tune pre-trained BERT
representations on an annotated dataset using the
BertForTokenClassification module from the HuggingFace library (Wolf et
al. 2019).

\hypertarget{replacing-identifiable-tokens}{%
\subsection{Replacing identifiable
tokens}\label{replacing-identifiable-tokens}}

Once we have identified PSI, we use a rule-based approach to replace
them with generic terms. To do this, we enumerate instances of a
specific class and replace all occurrences of that instance in the text
with a generic identifier representing the class to which the sensitive
phrase belongs. For example, assuming we identify the word ``London'' as
``LOCATION'' in the text, we replace it with the term ``LOCATION\_N.''
Here, N is a number that uniquely identifies ``London'' from all other
identified LOCATION phrases and ensures that subsequent mentions of
London in the same document are also mapped to the identical replacement
(e.g., multiple mentions of London in one document all become
LOCATION\_1).

\begin{figure}

{\centering \includegraphics{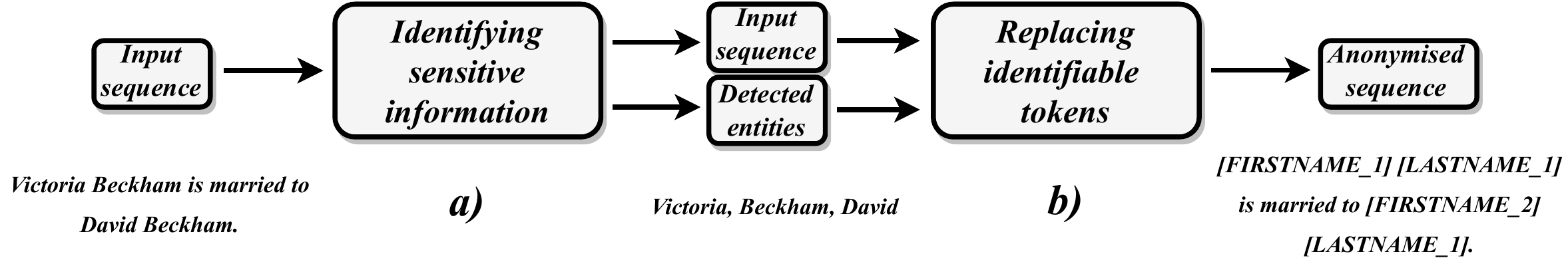} 

}

\caption{Illustration of the end-to-end anonymisation process of Textwash.}\label{fig:viz/textwash.pdf}
\end{figure}

Figure 1 shows how, given the input sequence ``Victoria Beckham is
married to David Beckham,'' Textwash identifies sensitive information
(a). The tool then inputs the original sequence and the detected
entities into the second module, which replaces the identifiable tokens
in a meaning-preserving way (b).

\hypertarget{dataset}{%
\subsection{Dataset}\label{dataset}}

The dataset used to train Textwash consists of 3,717 articles where each
phrase (i.e., single or multiple words in a sequence) is annotated
according to whether it represents PSI. It contains 417 articles from
the British National Corpus (BNC; Consortium and others (2007)), 1,800
emails (email body only) from the Enron email dataset
(\url{https://www.cs.cmu.edu/~enron/}) and 1,500 Wikipedia articles. For
the latter, we randomly sampled Wikipedia articles about persons that
contain at least 100 words. We excluded documents with less than 20
words from voth datasets. For the Enron dataset, we furthermore excluded
documents with more than 500 words, and for the BNC, we truncated all
documents to the first 500 words. We then sampled documents from both
datasets according to the highest named entity ratios.\footnote{The
  named entity ratio was the number of named entities - irrespective of
  their category - divided by the number of words in the document.}
However, for Enron, we only select documents with a named entity ratio
of less than 20\% since we observed that otherwise a substantial
proportion of emails consist of simple lists of names rather than
continuous text. Choosing a high named entity ratio ensured that the
documents were sufficiently rich in PSI to allow the machine learning
model to learn to identify its various categories.

Two domain experts annotated the dataset with the following tags:

\begin{itemize}
\tightlist
\item
  PERSON\_FIRSTNAME: a person's first name (e.g., Jane)
\item
  PERSON\_LASTNAME: a person's last name (e.g., Doe)
\item
  OCCUPATION: an occupation (e.g., nurse, carpenter)
\item
  LOCATION: a location (e.g., London, Berlin, France)
\item
  TIME: a time (e.g., 12 pm, afternoon)
\item
  ORGANIZATION: an organisation (e.g., Google, NHS)
\item
  DATE: a reference to a specific day (e.g., 12/10/2021, yesterday)
\item
  ADDRESS: an address (e.g., 42 London Road)
\item
  PHONE\_NUMBER: a phone number
\item
  EMAIL\_ADDRESS: an email address
\item
  OTHER\_IDENTIFYING\_ATTRIBUTE: an identifying attribute that cannot be
  categorised into the above but is still considered PSI
\item
  NONE: all other tokens in the input sequence
\end{itemize}

To train Textwash, we randomly split the dataset into a training,
validation and test set by using 80\% for training and 10\% each for
validation and testing.

The following sections detail two studies on the empirical validation of
the Textwash tool, including automated performance metrics and human
benchmarking.

\hypertarget{study-1}{%
\section{Study 1}\label{study-1}}

\hypertarget{method}{%
\subsection{Method}\label{method}}

We evaluate Textwash using the TILD criteria (Mozes and Kleinberg 2021).
These go beyond a technical evaluation (i.e., how many entities are
correctly identified per category) and include a test of the information
loss and a human de-anonymisation test. Information loss refers to the
difference in text analysis outcomes attributable to the anonymisation
procedure and is further divided into utility loss (i.e., the difference
between original and anonymised texts in prediction tasks) and construct
loss (i.e., the difference in linguistic variables or text statistics
between original and anonymised version). On the other hand, the
de-anonymisation evaluation examines whether an individual's identity is
leaking from anonymised text data. Since that test is hard to conduct
automatically, we adopt the `motivated intruder' principle and task
human participants with the de-anonymisation of text data.

\hypertarget{technical-evaluation}{%
\subsubsection{Technical evaluation}\label{technical-evaluation}}

Desirable for a good anonymisation tool is a high detection accuracy on
the sub-categories of phrases that are identified for replacement in a
subsequent step (here: first names, locations, etc.). We report the
detection results of anonymisation categories on the test set (i.e.,
unseen data). Each token in the test set's input sequences is labelled
according to the aforementioned categories, and we assess the model's
performance in predicting the respective label for each word.
Specifically, we report the precision, recall and F1-score for each
category, as well as macro and weighted averages.\footnote{The macro
  accuracy expresses the average irrespective of each class's support
  (i.e., how many occurrences does this category have in the test set?),
  whereas the weighted accuracy weighs the score for each category based
  on its support.}

\hypertarget{information-loss}{%
\subsubsection{Information loss}\label{information-loss}}

The two sub-categories of information loss are utility loss (i.e., the
difference in some performance metric on a classification task between
original and anonymised data) and construct loss (i.e., the difference
in variables between original and anonymised data). Higher loss values
imply a decreased performance (utility loss) or variable value
(construct loss) in the anonymised text compared to its original
counterparts. Small loss values are desirable and would suggest that the
anonymisation procedure retains the usefulness of the text data for
downstream analyses.

\hypertarget{utility-loss}{%
\paragraph{Utility loss}\label{utility-loss}}

For the utility loss, we investigated to what extent the anonymisation
of text data affected text classification performance. We used the IMDb
movie reviews dataset (Maas et al. 2011), a dataset widely used in NLP
research for sentiment analysis. The dataset consists of 50,000 movie
reviews, each annotated with a positive or negative sentiment label. For
training and testing, the dataset has a pre-defined split of 25,000
samples (balanced across sentiment).

RoBERTa is a neural network-based machine learning model based on the
Transformer architecture (Vaswani et al. 2017), widely used in NLP
research (Xia, Wu, and Van Durme 2020). The model has been pre-trained
on large corpora of text in a self-supervised fashion by learning to
predict masked tokens in textual input sequences. At the model's core
lies an attention mechanism capable of learning to capture words in
different contexts. The pre-trained layers of the RoBERTa model are then
fine-tuned for a specific downstream task, which is classification in
our case. To evaluate the utility loss in classification results, we
first used the original, non-anonymised data and fine-tuned a pretrained
RoBERTa model (Liu et al. 2019) on the training set, holding out 1,000
randomly selected training set sequences for validation. After training,
we tested the model on the test set to measure the performance on unseen
data. We report the model accuracy in predicting the correct sentiment
label for each test set sequence.

We then anonymised the entire dataset using Textwash and repeated the
procedure from above (training, validation, testing). The utility loss
is the difference in model prediction accuracy on the test set before
and after anonymisation.

\hypertarget{construct-loss}{%
\paragraph{Construct loss}\label{construct-loss}}

Since researchers may also be interested in linguistic variables, we
tested how much the values obtained from anonymised text data deviated
from those derived from the original text data. The construct loss was
assessed by testing whether the frequencies of part-of-speech tags
differed between original and anonymised texts\footnote{The frequency of
  part-of-speech tags was extracted with the \emph{spacyr} package
  (Benoit and Matsuo 2017) using the Universal POS tags from
  \url{https://universaldependencies.org/u/pos/}}. Since we wish to
quantify statistical evidence of the absence of a difference (i.e.,
supporting the assertion that both texts result in the same values), we
use Bayesian hypothesis testing for this (Rouder et al. 2009)\footnote{Bayesian
  t-tests allow us to quantify evidence for the null hypothesis (i.e.,
  original = anonymised) as well as for an alternative hypothesis (i.e.,
  original \textgreater{} anonymised).} We report the Bayes factor in
either direction: \(BF_{01}\) for evidence for the null and \(BF_{10}\)
for evidence for the alternative hypothesis.

\hypertarget{de-anonymisation}{%
\subsubsection{De-anonymisation}\label{de-anonymisation}}

As the litmus test for any anonymisation tool, we investigated how well
human participants can identify individuals from anonymised text data.
In these terms, a tool's good performance corresponds to low
re-identification rates. From a data protection point of view, the
ability to sufficiently anonymise data is essential for the tool's
usefulness and supersedes the other criteria. We operationalised this
assessment through a motivated intruder test.

We first instructed a group of participants to write descriptions of
persons similar in style to the introductory paragraphs of a Wikipedia
entry.\footnote{We did not use actual Wikipedia entries because the
  sentence structure - even with anonymisation - could be used in search
  engines and quickly lead to the source text. This would not represent
  de-anonymisation in a meaningful sense.} In order to submit the
anonymisation tool to tests of varying degrees of difficulty, we
elicited person descriptions of very famous, semifamous and non-existing
individuals. The rationale behind that decision was to provide a
complete picture of what the tool can achieve in different contexts.

Very famous individuals are the most challenging test for an
anonymisation tool because de-anonymisation can often hinge on specific
and seemingly-innocuous details that reveal an identity (i.e., the range
of PSI is exceptionally broad). For example - and as shown further below
- the UK singer Adele could be easily identified through mentions of
``singer'' (for which she is famous) and ``weight loss'' (which was
widely discussed in her case). Notably, the identification of Adele in
this case only works because this privileged information is in the
public domain and a human de-anonymiser could combine these pieces of
information to make an educated guess. As such, very famous persons may
be identifiable through properties that no longer meet the definitions
of anonymisation as set out elsewhere (Information Commissioner's Office
2012) and are somewhat removed from research use cases (e.g.,
anonymising interview data from participants who are not widely known to
the general public). Thus, very famous people present the lower bounds
of what the tool can achieve (i.e., less famous persons would be more
anonymisable).

At the other end of the difficulty spectrum, we use person descriptions
for individuals who do not exist. These persons cannot be Googled or
searched for in any other capacity, so identifying them by name is only
possible if the name itself has leaked. In between these two extremes
(very famous celebrities and non-existing persons), we use
``semifamous'' individuals who all have a Wikipedia entry but are not
well known generally. For applications of the Textwash tool on research
or business data, the performance on semifamous and non-existing persons
is thus the most meaningful benchmark.

\hypertarget{gathering-person-descriptions}{%
\paragraph{Gathering person
descriptions}\label{gathering-person-descriptions}}

\begin{itemize}
\tightlist
\item
  Task: After providing informed consent, the participants were
  instructed to write a person description of at least 500 characters
  for three individuals. They wrote (in that order) one description of a
  very famous person, one of a semifamous person, and one of a
  non-existing, fictitious person. The person subjects were chosen at
  random from a pre-defined item pool. The data collection was done in
  Qualtrics.
\item
  Item pool: The following items were used as pointers for the
  participants - each category consisted of 10 items:

  \begin{itemize}
  \tightlist
  \item
    Very famous persons: the participants were presented with the name
    of a very well-known person in the UK (e.g., Emma Watson, Benedict
    Cumberbatch).
  \item
    Semifamous persons: the participants were presented with the name of
    a lesser-known person in the UK (e.g., Irvin Brooks, Kenny Kramm)
    and encouraged to look up additional information on Wikipedia. The
    URL to the person's Wikipedia page was provided.
  \item
    Fictitious persons: we used the \emph{charlatan} R package
    (Chamberlain and Voytovich 2020) to generate profiles consisting of
    a full name (e.g., Amelie Crooks), a relationship status (e.g.,
    married), a job (e.g., software engineer), an age and a country of
    residence (e.g., Belgium).
  \end{itemize}
\item
  Participants: A total of \(n=401\) participants wrote three person
  descriptions each, resulting in a corpus of 1202 person descriptions
  (famous: 401, semifamous: 401, fictitious: 400)\footnote{Note that one
    description of a fictitious person was missing.}. The participants
  who wrote these person descriptions had a mean age of 34.79 years
  (\(SD=11.99\)), with 65.58\% females. The data were collected through
  Prolific Academic and each participant was paid GBP 1.25 per set of
  three descriptions.
\item
  Corpus details: The person descriptions were, on average, 130.49
  tokens long (\(SD=35.17\)), which did not differ between the type of
  person description, \(F(2, 1199) = 2.25, p=.106\),
  (\(M_{famous} = 128.54\), \(SD_{famous} = 29.07\);
  \(M_{semifamous} = 129.44\), \(SD_{semifamous} = 41.94\);
  \(M_{fictitious} = 133.48\), \(SD_{fictitious} = 33.16\)). The
  paragraph below shows a verbatim example of a description from the
  `very famous' category (person described: actress Emma Watson).
\end{itemize}

\emph{Emma Watson is most well known for starring as Hermione Granger in
Harry Potter, she was in all of the films. She is also known for playing
Belle in beauty and the beast. She has strong feminist views. Emma
Watson was born in Paris and bought up in Oxfordshire. Emma Watson is
currently 30 years old, her date of birth being 15th April 1990. Emma
Watson has 4 siblings, she attended Brown University. Both Emma
Watsons's parents are lawyers. Emma Watson speaks French but claims not
as well as she used to. Emma Watson has also ventured into modelling,
she became the face of Burberry and resulted in her earning a 6 figure
sum from the campaign.}

\hypertarget{motivated-intruder-testing}{%
\paragraph{Motivated intruder
testing}\label{motivated-intruder-testing}}

\begin{itemize}
\tightlist
\item
  Task: We created three tasks, one for each difficulty level (famous,
  semifamous, fictitious). Each participant in the motivated intruder
  task judged a random selection of ten person descriptions in the
  respective group, and we aimed to collect three motivated intruder
  judgments for each person description. After providing informed
  consent, the participants were informed that they had to ``play an
  adversary'' whose task it was to find out who a given person's
  description was about. After reading a description, they had to state
  i) whether they could identify the person (yes/no), ii) who they think
  it is (if they could identify the person) or the extent of knowledge
  that they could establish about the person (if they could not identify
  the person), and iii) which part of the text revealed the identity.
  After making these judgments for ten person descriptions, each
  participant was debriefed, automatically reimbursed for their time
  (1.50 GBP) and redirected to Prolific Academic. The data collection
  task was conducted in a custom-made web interface.
\item
  Participants: A total of \(n=366\) participants completed this part of
  the study. Of these participants, \(n=122\) judged descriptions of
  famous persons, \(n=123\) those of semifamous persons, and \(n=121\)
  those of fictitious persons. The mean age of participants was 27.31
  years (\(SD=9.37\)), with 40.98\% female participants. In total, the
  dataset consisted of 3660 judgments. Each person description was
  judged, on average, 3.06 times (\(SD=1.09\)).
\item
  Measuring de-anonymisation: We examined the de-anonymisation of
  individuals by participants by calculating the string similarity
  between the correct solution (e.g., Sam Smith) and the
  participant-indicated person. Specifically, we calculated the cosine
  similarity between the vectors of individual characters (e.g., s, a,
  m, s, m, i, t, h). The resulting similarity is a score between -1.00
  and +1.00, with values closer to +1.00 indicating higher similarity.
  The multiple judgments per item were averaged to obtain a single
  (average) similarity per item. To allow for fuzziness (e.g., typos),
  we chose a cosine similarity of 0.75 as the cut-off above which we
  deemed a person successfully identified, resulting in a binary outcome
  (identified vs not identified). In addition to that binarisation, we
  used cosine similarity as a continuous variable to assess the degree
  of similarity.
\end{itemize}

\hypertarget{results}{%
\subsection{Results}\label{results}}

\hypertarget{technical-evaluation-1}{%
\subsubsection{Technical evaluation}\label{technical-evaluation-1}}

When assessing the tagging performance of the entity-detection model, we
observe that the model accurately predicts most categories, with
F1-scores above 80\% for 10 out of 12 categories (Table 1). However, we
also observe that the model performs poorly on OCCUPATION (F1-score of
52\%) and OTHER\_IDENTIFYING\_ATTRIBUTE (F1-score of 69\%). To be clear,
a confusion of categories does not necessarily mean that a phrase is not
removed during anonymisation; just that it may not have been replaced
with the correct context-preserving placeholder. The high F1-score for
the NONE category (96\%) implies that sub-categories notwithstanding,
PSI is successfully identified.

\begin{longtable}[]{@{}lllll@{}}
\caption{Performance results of the trained Textwash model on the test
set.}\tabularnewline
\toprule
Entity tag & Precision & Recall & F1-score & Support \\
\midrule
\endfirsthead
\toprule
Entity tag & Precision & Recall & F1-score & Support \\
\midrule
\endhead
ADDRESS & 0.84 & 0.93 & 0.88 & 556 \\
DATE & 0.95 & 0.95 & 0.95 & 3301 \\
EMAIL\_ADDRESS & 0.96 & 0.98 & 0.97 & 1815 \\
LOCATION & 0.77 & 0.83 & 0.80 & 2111 \\
OCCUPATION & 0.43 & 0.65 & 0.52 & 307 \\
ORGANIZATION & 0.85 & 0.79 & 0.82 & 7300 \\
PERSON\_FIRSTNAME & 0.91 & 0.89 & 0.90 & 4278 \\
PERSON\_LASTNAME & 0.94 & 0.91 & 0.92 & 6434 \\
PHONE\_NUMBER & 0.96 & 0.95 & 0.95 & 874 \\
TIME & 0.90 & 0.91 & 0.90 & 934 \\
OTHER\_IDENTIFYING\_ATTRIBUTE & 0.64 & 0.74 & 0.69 & 3292 \\
NONE & 0.97 & 0.96 & 0.96 & 55134 \\
--------------------------- & --------- & ------ & -------- & ------- \\
accuracy & & & 0.93 & 86336 \\
macro avg & 0.84 & 0.87 & 0.86 & 86336 \\
weighted avg & 0.93 & 0.93 & 0.93 & 86336 \\
\bottomrule
\end{longtable}

\hypertarget{information-loss-1}{%
\subsubsection{Information loss}\label{information-loss-1}}

\hypertarget{utility-loss-1}{%
\paragraph{Utility loss}\label{utility-loss-1}}

On the original dataset, the model achieves a test set accuracy of
92.82\% (precision: 92.69\%, recall: 92.97\%, F1: 92.83\%), which is
comparable to existing work (Mozes et al. 2021). On the anonymised
dataset, the trained RoBERTa model achieves an accuracy of 91.98\%
(precision: 91.41\%, recall: 92.67\%, F1: 92.04\%) on the test set,
resulting in a utility loss of 0.84\%. These findings indicate that
anonymising documents has a negligible influence on the performance of
the sentiment classification task, thereby retaining the usefulness of
the anonymised dataset for downstream text classification tasks.

\hypertarget{construct-loss-1}{%
\paragraph{Construct loss}\label{construct-loss-1}}

The findings for the construct loss analysis indicate that the
information is preserved for the POS tags where that was expected (Table
2). We see substantial deviations from the original text data for
adjectives, adverbs, nouns, numerals, pronouns and proper nouns - all of
these are detected in the Textwash tool and replaced with placeholders
that are not captured by POS tagging algorithms. If one were to map the
replacements (e.g., PERSON\_X) to desired POS tags (here: noun), the POS
information would be fully recoverable. As a whole, these findings
suggest that anonymisation retains the usefulness of linguistic
variables.

\begin{longtable}[]{@{}llll@{}}
\caption{Bayes factors for the construct loss test on POS frequencies
(original vs anonymised text data).}\tabularnewline
\toprule
POS & Description & \(BF_{10}\) & \(BF_{01}\) \\
\midrule
\endfirsthead
\toprule
POS & Description & \(BF_{10}\) & \(BF_{01}\) \\
\midrule
\endhead
X & Other & 0.05 & 21.03 \\
INTJ & Interjections & 0.05 & 20.85 \\
CCONJ & Coordinating conjunction & 0.05 & 19.72 \\
DET & Determiners & 0.05 & 19.66 \\
VERB & Verbs & 0.06 & 16.26 \\
SYM & Symbols & 0.06 & 16.17 \\
PART & Particles & 0.07 & 13.71 \\
ADP & Adposition & 0.09 & 11.66 \\
ntok & No.~of words & 0.19 & 5.17 \\
PUNCT & Punctuation & 0.20 & 4.98 \\
ADJ & Adjectives & 1.277946e+35 & 0.00 \\
ADV & Adverbs & 5.712415e+05 & 0.00 \\
NOUN & Nouns & 5.369620e+182 & 0.00 \\
NUM & Numerals & 1.192332e+223 & 0.00 \\
PRON & Pronouns & Inf & 0.00 \\
PROPN & Proper nouns & 1.984156e+28 & 0.00 \\
\bottomrule
\end{longtable}

\hypertarget{de-anonymisation-1}{%
\subsubsection{De-anonymisation}\label{de-anonymisation-1}}

\hypertarget{overall-de-anonymisation}{%
\paragraph{Overall de-anonymisation}\label{overall-de-anonymisation}}

Table 3 shows the de-anonymisation rates for each person description
level.\footnote{The participants' self-reported success in
  de-anonymisation - ``Could you identify the person?'' - had high
  agreement with the actual de-anonymisation of 92.43\%. We use the
  actual de-anonymisation as a criterion throughout the paper.} As
expected, very famous people are identified more often than semifamous
or fictitious people. Almost 19\% of famous persons are identified in
the motivated intruder test, while merely 2\% and 1\% of the semifamous
and fictitious persons could be identified by name, respectively.

\begin{longtable}[]{@{}lllll@{}}
\caption{Cosine similarities between the true person name and the
participant choice (M, SD) and (un)successful de-anonymisations per
type.}\tabularnewline
\toprule
Item type & \(M\) & \(SD\) & \% identified & SE \% identified \\
\midrule
\endfirsthead
\toprule
Item type & \(M\) & \(SD\) & \% identified & SE \% identified \\
\midrule
\endhead
famous & 0.41 & 0.36 & 18.25 & 1.93 \\
fictitious & 0.04 & 0.13 & 1.01 & 0.50 \\
semifamous & 0.13 & 0.20 & 2.01 & 0.70 \\
\bottomrule
\end{longtable}

\hypertarget{person-analysis-and-information-leakage}{%
\paragraph{Person analysis and information
leakage}\label{person-analysis-and-information-leakage}}

There was considerable variation among the items in the de-anonymisation
rate and cosine string similarity. To avoid flooring effects, we only
looked at the famous persons' descriptions (Table 4). While some famous
persons were only identified in less than 10\% of the cases
(Cumberbatch, Jagger, Bale), some had a de-anonymisation rate of over
25\% (John, Radcliffe, Smith).

\begin{longtable}[]{@{}lllll@{}}
\caption{Mean (SD) cosine similarities per famous person and
de-anonymisation rate (\%) with standard error (SE) for Study
1.}\tabularnewline
\toprule
Name & \(M\) & \(SD\) & \% identified & SE \% identified \\
\midrule
\endfirsthead
\toprule
Name & \(M\) & \(SD\) & \% identified & SE \% identified \\
\midrule
\endhead
benedict cumberbatch & 0.25 & 0.29 & 4.65 & 3.25 \\
sam smith & 0.50 & 0.40 & 32.50 & 7.50 \\
ed sheeran & 0.56 & 0.34 & 28.95 & 7.46 \\
emma watson & 0.39 & 0.34 & 16.67 & 5.82 \\
elton john & 0.47 & 0.37 & 25.00 & 6.60 \\
mick jagger & 0.30 & 0.31 & 7.32 & 4.12 \\
adele & 0.46 & 0.35 & 18.42 & 6.37 \\
daniel radcliffe & 0.48 & 0.37 & 30.00 & 7.34 \\
christian bale & 0.36 & 0.34 & 8.33 & 4.67 \\
hugh grant & 0.29 & 0.32 & 10.53 & 5.05 \\
\bottomrule
\end{longtable}

To understand how participants identified these individuals, we looked
at the information provided in the motivated intruder test when asked
``what revealed the identity for you?'' We analysed the answers that
participants provided when they were successful by looking at the
n-grams\footnote{An n-gram is a sequence of \emph{n} tokens. We removed
  stopwords before obtaining n-grams.} (unigrams, bigrams, trigrams)
most telling for each item. Table 5 shows the top 10 n-grams per item.
We see that the persons that had relatively high de-anonymisation rates
were identified through the leakage of very specific details:
``glasses'' and ``gay'' were mentioned for Elton John and presumably
allowed participants to combine information about the person being a
singer and these attributes in a ``guesstimate'' that it might be Elton
John. Daniel Radcliffe was identified through being an actor playing a
role in Harry Potter, and for Sam Smith, the participants mentioned song
titles and ``gay'' as revealing attributes.

\begin{longtable}[]{@{}
  >{\raggedright\arraybackslash}p{(\columnwidth - 2\tabcolsep) * \real{0.16}}
  >{\raggedright\arraybackslash}p{(\columnwidth - 2\tabcolsep) * \real{0.84}}@{}}
\caption{Top-10 n-grams that revealed the person's identity in the
motivated intruder test in Study 1.}\tabularnewline
\toprule
Name & Top-10 n-grams \\
\midrule
\endfirsthead
\toprule
Name & Top-10 n-grams \\
\midrule
\endhead
adele & weight, song, songs, loss, weight\_loss, names, famous, fact,
titles, name \\
benedict cumberbatch & strange, doctor, doctor\_strange, fact, roles,
sherlock, movie, played, role, description \\
christian bale & batman, role, movie, movies, american, psycho, oscar,
dark, american\_psycho, film \\
daniel radcliffe & harry, potter, harry\_potter, young, black, boy, j.k,
actor, woman, woman\_black \\
ed sheeran & hair, ginger, songs, ginger\_hair, red, name, musician,
names, music, red\_hair \\
elton john & songs, song, singer, glasses, gay, name, music, names,
first, description \\
emma watson & harry, potter, harry\_potter, beauty, beast,
beauty\_beast, role, actress, feminist, activist \\
hugh grant & movies, bridget, jones, bridget\_jones, film, name, hill,
movie, actor, played \\
mick jagger & rolling, stones, rolling\_stones, singer, song, name,
singer\_rolling, singer\_rolling\_stones, band, fact \\
sam smith & song, songs, name, title, la, wall, singer, song\_title,
name\_song, gay \\
\bottomrule
\end{longtable}

\hypertarget{non-identifying-information-leakage}{%
\paragraph{Non-identifying information
leakage}\label{non-identifying-information-leakage}}

In addition to the \emph{revealing leakage} that helped intruders
identify a person, we also asked those who did not identify the person
what they could still recover as information from the anonymised text
data. Table 6 shows the most mentioned n-grams per person and suggests
that in cases \emph{when the person was not identified}, it was mainly
generic information. It is possible that in some cases, the intruder was
not able to combine these pieces effectively: for example, Ed Sheeran
could have been identified with a combination of the attributes
``ginger,'' ``hair,'' ``pop,'' and ``music.''

\begin{longtable}[]{@{}
  >{\raggedright\arraybackslash}p{(\columnwidth - 2\tabcolsep) * \real{0.20}}
  >{\raggedright\arraybackslash}p{(\columnwidth - 2\tabcolsep) * \real{0.80}}@{}}
\caption{Top-10 n-grams that were leaked but did not allow the motivated
intruder to identify the person in Study 1.}\tabularnewline
\toprule
Name & Top-10 n-grams \\
\midrule
\endfirsthead
\toprule
Name & Top-10 n-grams \\
\midrule
\endhead
adele & singer, famous, person, female, musician, know, music, pop,
awards, famous\_singer \\
benedict cumberbatch & actor, person, famous, know, movies, male, movie,
married, tv, man \\
christian bale & actor, person, male, movies, famous, nothing, won,
roles, many, movie \\
daniel radcliffe & actor, person, hair, know, famous, films, brown,
brown\_hair, movie, male \\
ed sheeran & ginger, know, singer, hair, daughter, ginger\_hair, person,
music, pop, lot \\
elton john & singer, famous, gay, person, musician, married, song,
songs, children, name \\
emma watson & actress, model, actor, person, female, feminist, famous,
know, actress\_model, nothing \\
hugh grant & actor, film, producer, romantic, movies, person, plays,
know, english, one \\
mick jagger & person, singer, band, famous, know, rock, married, male,
well, musician \\
sam smith & singer, gay, person, songwriter, won, voice, non-binary,
musician, childhood, lot \\
\bottomrule
\end{longtable}

\hypertarget{item-level-analysis}{%
\paragraph{Item-level analysis}\label{item-level-analysis}}

The findings presented above are aggregations of items per person
described. But since each person has been described in multiple texts,
there may be variation in the de-anonymisability even across items
relating to the same individual. On average, each person has been
described in 40 unique person descriptions (\(SD=2.54\)), of which each
was subject to the motivated intruder test on average 3.06 times
(\(SD=1.09\)). We explored how - for one person - the de-anonymisability
differed. Below we show examples for two persons (Ed Sheeran and Sam
Smith) - one for each that was always identified and one that was never
identified.

\begin{itemize}
\tightlist
\item
  Ed Sheeran (never identified): \emph{PERSON\_FIRSTNAME\_2
  PERSON\_LASTNAME\_1 is a famous LOCATION\_1 musician - singer and
  songwriter. Born on DATE\_4 in Halifax, UK, under the full name
  PERSON\_FIRSTNAME\_1 PERSON\_LASTNAME\_1 . PRONOUN estimated net worth
  as of DATE\_3 is NUMERIC\_5 NUMERIC\_4 NUMERIC\_4. PRONOUN is also
  known as a record producer, as well as actor PRONOUN played PRONOUN in
  a LOCATION\_4 soap opera OTHER\_IDENTIFYING\_ATTRIBUTE\_1, which was
  filmed while PRONOUN was in the country in DATE\_2 for a
  NUMERIC\_1-off performance). PRONOUN career began somewhere in
  DATE\_5, but PRONOUN shot to international fame in DATE\_1. and had an
  immense success ever since. Currently, PRONOUN is said to be the
  NUMERIC\_7th richest musician in the LOCATION\_2.}
\item
  Ed Sheeran (always identified): \emph{PERSON\_FIRSTNAME\_2 is a famous
  musician,singer and songwritter. PRONOUN has ginger hair and wears
  glasses. PRONOUN has just had a baby i think. PRONOUN is english and
  has written songs for lots of other artists PRONOUN songs are very
  popular all around the world PRONOUN wrote a really good song PRONOUN
  did with OTHER\_IDENTIFYING\_ATTRIBUTE\_1 swift.i like PRONOUN album
  shape of you . its really good PRONOUN is also a record producer and
  an actor PRONOUN has sold more than NUMERIC\_3million records
  worldwide PRONOUN has won all sorts of awards e.g.honoary degrees from
  universities,the MbEFOR SERVICES TO MUSIC PRONOUN lives in
  LOCATION\_1. i really dont know what else to write. im sorry but that
  is all i know. it pretty much covers everything you would want to know
  about PERSON\_FIRSTNAME\_2.}
\item
  Sam Smith (never identified): \emph{PERSON\_FIRSTNAME\_4
  PERSON\_LASTNAME\_3 is an internationally successful
  singer-songwriter. They were born on DATE\_2 in LOCATION\_1. They
  became famous in DATE\_1 and have been nominated and won multiple
  musical awards. PERSON\_FIRSTNAME\_6 announced NUMERIC\_1 years ago
  that they were genderqueer and preferred the pronoun, `they.'
  PERSON\_FIRSTNAME\_6 has recorded NUMERIC\_2 albums:
  OTHER\_IDENTIFYING\_ATTRIBUTE\_1 and OTHER\_IDENTIFYING\_ATTRIBUTE\_2.
  PERSON\_FIRSTNAME\_6's networth is estimated at NUMERIC\_4 million.
  Their main musical genres are R\&B, pop and soul and they are signed
  for ORGANIZATION\_1. Famous relatives include their third cousins:
  singer PERSON\_FIRSTNAME\_3 PERSON\_LASTNAME\_1 and actor
  PERSON\_FIRSTNAME\_5 PERSON\_LASTNAME\_1. PERSON\_FIRSTNAME\_6 is gay
  and has previously dated actor and model PERSON\_FIRSTNAME\_1
  PERSON\_LASTNAME\_4 and actor PERSON\_FIRSTNAME\_2
  PERSON\_LASTNAME\_2.}
\item
  Sam Smith (always identified): \emph{PERSON\_FIRSTNAME\_1
  PERSON\_LASTNAME\_2 was born on DATE\_2. PRONOUN is a
  singer/songwriter. PRONOUN was born in LOCATION\_1 and has net worth
  of NUMERIC\_4 million . PRONOUN hits include Money on my Mind,
  Writings on the Wall and OTHER\_IDENTIFYING\_ATTRIBUTE\_3. PRONOUN
  debut album was OTHER\_IDENTIFYING\_ATTRIBUTE\_2. PRONOUN has won
  several awards including NUMERIC\_5 Grammy Awards and a Golden Globe
  award. PRONOUN parents are called PERSON\_FIRSTNAME\_2
  PERSON\_LASTNAME\_2 and PERSON\_FIRSTNAME\_3 PERSON\_LASTNAME\_1.
  PRONOUN had liposuction when PRONOUN only NUMERIC\_3 years old and
  reported to be bullied badly as a child. PRONOUN came out as gay in
  DATE\_3 and then said PRONOUN was genderqueer in DATE\_1. PRONOUN is
  quoted as saying I feel as much a woman as a man.}
\end{itemize}

These examples offer a glimpse at the difficulty in text anonymisation
attributable to tiny details that, when combined, reveal an identity. In
Study 2, we seek to learn more about identifiable and non-identifiable
text data properties.

\hypertarget{discussion}{%
\subsection{Discussion}\label{discussion}}

Study 1 shows that i) the Textwash anonymisation can anonymise
practically all person descriptions of individuals who are not famous
and about whom intruders did not, therefore, have access to ``privileged
information.'' The most challenging test (anonymising the most famous
individuals in the UK) was successful in 81.75\% of the cases. We
further found that ii) anonymisation is very hard if even minute details
are leaked. In the case of very famous persons, a song title, reference
to recent weight loss (Adele) or movie role (Christine Bale, Benedict
Cumberbatch) can be sufficient for de-anonymisation. We also found that
iii) considerable variation in identifiability exists even in person
descriptions of the same individual.

To further improve the anonymisation, it would thus be helpful to
understand how identified text data differ from fully anonymised texts.
In Study 2, we extend and replicate the findings for the famous persons
group and examine the properties of (non-)identified person
descriptions.

\hypertarget{study-2}{%
\section{Study 2}\label{study-2}}

\hypertarget{method-1}{%
\subsection{Method}\label{method-1}}

Since the aim of Study 2 was to replicate findings from Study 1 and
understand the properties of unsuccessful and successful anonymisation,
we sought to avoid flooring effects and used only very famous
individuals. The list of individuals was extended to a total of 20
famous people. We gathered a new sample of person descriptions for all
of them.

\hypertarget{person-descriptions}{%
\subsubsection{Person descriptions}\label{person-descriptions}}

The task was identical to the person description task from Study 1 with
three exceptions: the item pool was extended to 20 person descriptions,
and each participant was paid GBP 3.75 for the task and wrote five
descriptions. A total of \(n=200\) participants (\(M_{age}=31.07\)
years, \(SD=8.32\), 73.00\% female) wrote 1080 person descriptions with
an average length of 112.13 tokens (\(SD=24.68\)). Each person was
described in - on average - 54 person descriptions (\(SD=4.09\)).

\hypertarget{motivated-intruder-test}{%
\subsubsection{Motivated intruder test}\label{motivated-intruder-test}}

We recruited \(n=222\) participants from Prolific Academic for the
motivated intruder task (\(M_{age} = 32.19\), \(SD=10.09\), 68.92\%
female). The task instructions were identical to those from Study 1.
Each participant was assigned a random selection of ten texts and was
paid GBP 1.75. We aimed to collect two judgments per text and obtained
2.06 judgments on average per item (\(SD=0.73\)).

\hypertarget{examining-text-anonymisability}{%
\subsubsection{Examining text
anonymisability}\label{examining-text-anonymisability}}

By choosing the most difficult text anonymisation task (i.e., very
famous people), we expect some texts to result in re-identifications of
the individual. That allows us to test whether texts that led to
de-anonymisation differ statistically from those that were not
de-anonymised on the following variables.

\hypertarget{proportion-of-anonymised-text}{%
\paragraph{Proportion of anonymised
text}\label{proportion-of-anonymised-text}}

We define the proportion of anonymised text as
\(P_{removed} = 1 - \frac{ntok_{anonymised}}{ntok_{original}}\), with
\(ntok\) being the number of tokens in the anonymised or original text.
We removed all anonymisation tags (e.g., {[}Person\_1{]}) from the
anonymised documents.

\hypertarget{global-frequency-ranks-of-anonymised-texts}{%
\paragraph{Global frequency ranks of anonymised
texts}\label{global-frequency-ranks-of-anonymised-texts}}

A look at the leaking information from Study 1 (Tables 5 and 6) suggests
that very specific pieces of information lead intruders to identify
individuals. We test whether the words in the de-anonymised documents
differ from those in the documents that did not result in
re-identification regarding their global occurrence frequency. The
frequency is operationalised as the average frequency rank of the words
in each document based on a list of the most frequent 10k words based on
the Google Trillion Word Corpus\footnote{\url{https://github.com/first20hours/google-10000-english}}.
Higher rank scores imply a low frequency. If highly specific - and hence
less frequent - words reveal an identity, we expect a higher global
frequency rank score for identified person descriptions than
non-identified ones.

\hypertarget{perplexity-of-anonymised-texts}{%
\paragraph{Perplexity of anonymised
texts}\label{perplexity-of-anonymised-texts}}

Another way to measure the ``unusualness'' of the information left in
the anonymised documents is perplexity. Perplexity is the inverse
exponentialised probability of observing a sequence of words as computed
by a language model. Low perplexity implies a higher probability
assigned to the input sequence and, therefore, a lower unusualness of
the text. We compute the perplexity of input text using a pre-trained
GPT language model (Radford et al. 2018) as provided by the HuggingFace
Transformers library (Wolf et al. 2019).

\hypertarget{results-1}{%
\subsection{Results}\label{results-1}}

\hypertarget{de-anonymisation-2}{%
\subsubsection[De-anonymisation]{\texorpdfstring{De-anonymisation\footnote{The
  agreement between participant-indicated identification and actual
  identification was 78.38\%. As in Study 1, we used the actual
  identification for analysis.}}{De-anonymisation}}\label{de-anonymisation-2}}

The mean de-anonymisation rate was 26.39\% (\(SE=1.34\)) with an average
cosine similarity of 0.42 (\(SD=0.39\)) between the actual person's name
and the participant's input. Table 7 shows the variation in
de-anonymisation among the items in the augmented stimuli set of Study
2. Some newly added person descriptions showed particularly high
de-anonymisation rates (e.g., Beckham, Hamilton, Middleton). When we
only consider the persons that were also among the famous people in
Study 1, we obtain a comparable de-anonymisation rate as in Study 1
(\(M=22.63\%\), \(SE=1.79\)).

\begin{longtable}[]{@{}lllll@{}}
\caption{Mean (SD) cosine similarities per famous person and
de-anonymisation rate (\%) with standard error (SE) for Study
2.}\tabularnewline
\toprule
Name & \(M\) & \(SD\) & \% identified & SE \% identified \\
\midrule
\endfirsthead
\toprule
Name & \(M\) & \(SD\) & \% identified & SE \% identified \\
\midrule
\endhead
adele & 0.53 & 0.41 & 38.98 & 6.40 \\
christian bale & 0.24 & 0.31 & 9.80 & 4.21 \\
david beckham & 0.68 & 0.39 & 55.10 & 7.18 \\
naomi campbell & 0.41 & 0.34 & 19.61 & 5.61 \\
daniel craig & 0.35 & 0.38 & 22.22 & 5.71 \\
benedict cumberbatch & 0.18 & 0.29 & 7.41 & 3.60 \\
cara delevigne & 0.36 & 0.40 & 23.53 & 6.00 \\
judi dench & 0.14 & 0.25 & 5.36 & 3.04 \\
ricky gervais & 0.47 & 0.39 & 28.26 & 6.71 \\
hugh grant & 0.39 & 0.40 & 22.92 & 6.13 \\
lewis hamilton & 0.69 & 0.35 & 50.82 & 6.45 \\
mick jagger & 0.26 & 0.32 & 9.43 & 4.05 \\
elton john & 0.57 & 0.37 & 38.89 & 6.70 \\
kate middleton & 0.60 & 0.34 & 44.64 & 6.70 \\
kate moss & 0.39 & 0.40 & 22.81 & 5.61 \\
daniel radcliffe & 0.25 & 0.33 & 10.17 & 3.97 \\
jk rowling & 0.49 & 0.35 & 29.41 & 6.44 \\
ed sheeran & 0.60 & 0.40 & 45.00 & 6.48 \\
sam smith & 0.40 & 0.35 & 16.07 & 4.95 \\
emma watson & 0.36 & 0.41 & 24.07 & 5.87 \\
\bottomrule
\end{longtable}

\hypertarget{information-leakage}{%
\subsubsection{Information leakage}\label{information-leakage}}

When we look at the leakage that led to successful identifications of
the ten most identified persons in Study 2, we observe that similar to
Study 1, very specific features revealed the identities of individuals
(Table 8). For example, Lewis Hamilton was identified by the combination
of ``formula 1'' and ``black,'' while David Beckham was identified
through the mention of being married to ``posh spice''\footnote{The name
  his wife, Victoria Beckham, used in the band Spice Girls.}.

\begin{longtable}[]{@{}
  >{\raggedright\arraybackslash}p{(\columnwidth - 2\tabcolsep) * \real{0.12}}
  >{\raggedright\arraybackslash}p{(\columnwidth - 2\tabcolsep) * \real{0.88}}@{}}
\caption{Top-10 n-grams that revealed the person's identity in the
motivated intruder test in Study 2.}\tabularnewline
\toprule
Name & Top-10 n-grams \\
\midrule
\endfirsthead
\toprule
Name & Top-10 n-grams \\
\midrule
\endhead
adele & song, weight, loss, weight\_loss, titles, songs, lost,
song\_titles, singer, age \\
cara delevigne & model, eyebrows, dyspraxia, book, singer, bisexual,
comic, comic\_book, actor, know \\
david beckham & spice, married, girl, spice\_girl, footballer, posh,
posh\_spice, married\_spice, married\_spice\_girl, football \\
ed sheeran & ginger, hair, song, guitar, singer, married, team,
ginger\_hair, names, album \\
elton john & pianist, song, gay, piano, singer, married, charity,
outfits, costumes, funeral \\
emma watson & harry, potter, harry\_potter, rights, actress, womens,
film, womens\_rights, feminist, activist \\
jk rowling & books, author, harry, series, potter, harry\_potter, book,
name, female, writer \\
kate middleton & royal, family, prince, throne, married, royal\_family,
line, line\_throne, married\_prince, future \\
lewis hamilton & driver, racing, formula, racing\_driver, 1, black, f1,
formula\_1, race, car \\
ricky gervais & office, comedian, comedy, shows, afterlife, reference,
life, actor, show, office\_afterlife \\
\bottomrule
\end{longtable}

\hypertarget{statistical-differences-in-text-anonymisability}{%
\subsubsection{Statistical differences in text
anonymisability}\label{statistical-differences-in-text-anonymisability}}

The comparisons between de-anonymised and anonymised documents on
linguistic variables (Table 9) suggest that the only effect observed was
for the proportion anonymised (effect size Cohen's \(d=0.19\)):
documents that were de-anonymised had a marginally lower percentage
removed from the original text than documents which remained anonymous.
The other variables (frequency ranks and perplexity) did not indicate a
significant difference.

\begin{longtable}[]{@{}llllll@{}}
\caption{Means (SDs) for anonymised (0) and de-anonymised documents (1)
for i) the proportion of anonymised text, global rank frequency, and
perplexity. The effect size Cohen's d (with 95\% CI) represents the
magnitude of the difference between anonymised and de-anonymised
documents per variable.}\tabularnewline
\toprule
Variable & \(M_0\) & \(SD_0\) & \(M_1\) & \(SD_1\) & \(d\) \\
\midrule
\endfirsthead
\toprule
Variable & \(M_0\) & \(SD_0\) & \(M_1\) & \(SD_1\) & \(d\) \\
\midrule
\endhead
Proportion anonymised & 22.94 & 8.77 & 21.45 & 7.09 & 0.19 {[}0.03;
0.35{]} \\
Frequency rank (original) & 1131.21 & 237.74 & 1109.18 & 241.19 & 0.09
{[}-0.07; 0.25{]} \\
Frequency rank (anonymised) & 1012.19 & 231.01 & 1003.55 & 224.06 & 0.04
{[}-0.12; 0.20{]} \\
Perplexity (original) & 66.68 & 47.96 & 69.29 & 35.98 & -0.06 {[}-0.23;
0.10{]} \\
Perplexity (anonymised) & 145.12 & 180.61 & 149.32 & 81.34 & -0.03
{[}-0.20; 0.13{]} \\
\bottomrule
\end{longtable}

The absence of evidence that successful anonymisation is detectable
through quantifiable linguistic indices suggests, again, that more
subtleties are at play. As a last exploratory analysis, we now look in
detail at three examples, their leaked information and the use thereof
by intruders for raw data.

\hypertarget{case-studies-to-understand-de-anonymisation}{%
\subsubsection{Case studies to understand
de-anonymisation}\label{case-studies-to-understand-de-anonymisation}}

The intruder task can be abstracted as follows: they read the anonymised
text, combine potentially leaking information to de-anonymise the text
and provide a brief explanation about what they used to de-anonymise the
individual. We look at three examples that shed light on that
decision-making in practice:

\textbf{Example 1: Kate Middleton}

\emph{Original text:} Kate Middleton is the wife of Prince William. She
is a mother of 3 children; 2 boys and a girl. Kate is educated to
university level and that is where she met her future husband. Kate
dresses elegantly and is often seen carrying out charity work. However,
she is a mum first and foremost and the interactions we see with her
children are adorable. Kate's sister, Pippa, has followed Kate into the
public eye. She was born in 1982 and will soon turn 40. When pregnant,
Kate suffers from a debilitating illness called Hyperemesis Gravidarum,
which was little known about until it was reported that Kate had it.

\emph{Anonymised text:} {[}firstname1{]} {[}lastname1{]} is the wife of
{[}occupation1{]} {[}lastname2{]}. {[}pronoun{]} is a mother of
{[}numeric{]} children; {[}numeric{]} boys and a girl. {[}firstname1{]}
is educated to university level and that is where {[}pronoun{]} met
{[}pronoun{]} future husband. {[}firstname1{]} dresses elegantly and is
often seen carrying out charity work. However, {[}pronoun{]} is a mum
first and foremost and the interactions we see with {[}pronoun{]}
children are adorable. {[}firstname1{]}'s sister, {[}firstname2{]}, has
followed {[}firstname1{]} into the public eye. {[}pronoun{]} was born in
{[}date1{]} and will soon turn {[}numeric{]}. When pregnant,
{[}firstname1{]} suffers from a debilitating illness called
{[}otherattribute1{]}, which was little known about until it was
reported that {[}firstname1{]} had it.

\emph{Information mentioned by intruder:} ``Suffered from an illness in
pregnancy and has a famous sister.''

\textbf{Example 2: Lewis Hamilton}

\emph{Original text:} Lewis Hamilton is a British racing driver. He
currently competes in Formula One for Mercedes, having previously driven
for McLaren. In Formula One, Hamilton has won a joint-record seven World
Drivers' Championship titles (tied with Michael Schumacher), and holds
the records for the most wins, pole positions, and podium finishes.
Hamilton is an advocate against racism and for increased diversity in
motorsport. More recently, he took the knee before every race he entered
in the 2020 Formula One season in support of the Black Lives Matter
movement and wore t-shirts bearing the Black Lives Matter slogan.
Following the murder of George Floyd, he criticised prominent figures in
Formula One for their silence on the issue.

\emph{Anonymised text:} {[}firstname1{]} {[}lastname1{]} is a
{[}location1{]} racing driver. {[}pronoun{]} currently competes in
{[}organisation1{]} for {[}organisation2{]}, having previously driven
for {[}organisation3{]}. In {[}organisation1{]}, {[}lastname1{]} has won
a joint-record {[}numeric{]} World Drivers' Championship titles (tied
with {[}firstname2{]} {[}lastname2{]} , and holds the records for the
most wins, pole positions, and podium finishes. {[}lastname1{]} is an
advocate against racism and for increased diversity in motorsport. More
recently, {[}pronoun{]} took the knee before every race {[}pronoun{]}
entered in the {[}date1{]} {[}organisation1{]} season in support of the
{[}otherattribute1{]} movement and wore t-shirts bearing the
{[}otherattribute1{]} slogan. Following the murder of {[}firstname3{]}
{[}lastname3{]}, {[}pronoun{]} criticised prominent figures in
{[}organisation1{]} for their silence on the issue.

\emph{Information mentioned by intruder:} ``A top racing driver.
advocate against racism and recently took the knee against racism. I
believe he criticised prominent people for their silence on the murder
of George Floyd.''

\textbf{Example 3: Daniel Craig}

\emph{Original text:} He is an English film actor known for playing
James Bond in the 007 series of films. Since 2005, he has been playing
the character but he confirmed that No Time to Die would be his last
James Bond film. He was born in Chester on 2nd of March in 1968. He
moved to Liverpool when his parents divorced and lived there until he
was sixteen years old. He auditioned and was accepted into the National
Youth Theatre and moved down to London. He studied at Guildhall School
of Music and Drama. He has appeared in many films.

\emph{Anonymised text:} {[}pronoun{]} is an {[}location1{]} film actor
known for playing {[}otherattribute1{]} in the {[}otherattribute2{]}
series of films. Since{[}date1{]}, {[}pronoun{]} has been playing the
character but {[}pronoun{]} confirmed that {[}otherattribute3{]} would
be {[}pronoun{]} last {[}otherattribute1{]} film. {[}pronoun{]} was born
in {[}location2{]} on {[}date2{]} of {[}date3{]} in {[}date4{]}.
{[}pronoun{]} moved to {[}location3{]} when {[}pronoun{]} parents
divorced and lived there until {[}pronoun{]} was {[}numeric{]} years
old. {[}pronoun{]} auditioned and was accepted into the
{[}organisation1{]} and moved down to {[}location4{]}. {[}pronoun{]}
studied at {[}organisation2{]}. {[}pronoun{]} has appeared in many
films.

\emph{Information mentioned by intruder:} ``The comment about it being a
last film.''

These examples illustrate the difficulty of anonymising text data. The
first example led to de-anonymisation through the mention of the person
having had an illness during pregnancy (not the actual name of the
illness, which was tagged and removed) and a famous sister. Lewis
Hamilton was identified by combining the information about being a
driver and having a firm stance against racism. At no place was the name
of the team he races for mentioned and references to the Black Lives
Matter movement or George Floyd were anonymised. Lastly, in the third
example, Daniel Craig was identified not through leakage of a name or
film title but through the note that ``{[}No Time To Die{]} would be
{[}his{]} last {[}James Bond{]} film.''

Note that in none of the cases was the name, a date of birth or
occupation leaked. Instead, it was ``soft'' identifiers that the
intruders could combine and connect to public knowledge to make well
educated guesses. All three cases highlight the importance of normally
``privileged information'': these individuals can be identified only
because very minute details about them are in the public domain, or
because the information included only applies to a very narrow range of
individuals.

\hypertarget{discussion-1}{%
\subsection{Discussion}\label{discussion-1}}

The second study replicated the findings from Study 1 for very famous
individuals. Only negligible differences in text variables emerged
between successfully and unsuccessfully anonymised texts on the
proportion of tagged tokens. A qualitative analysis of the
decision-making process suggests that minute details, combined with
public information, allow for the identification of famous persons.
Overall, the data show that text anonymisation works well and does not
leak information that - without external public knowledge - can lead to
the identification of an individual.

\hypertarget{general-discussion}{%
\section{General discussion}\label{general-discussion}}

Text anonymisation is one of the current hurdles in moving advances in
text analysis and NLP research closer to practice and enabling sharing
of text data. With an increased scale of datasets available and needed
for computational text analysis, automated text anonymisation tools
offer a promising approach to solving these issues. In this paper, we
introduced and validated Textwash - an open-source software that uses
machine learning and natural language processing to identify potentially
sensitive information in unstructured text data and remove it.
Importantly, the tool works with two end-users in mind: (1) for
researchers, the removal of information must be done so that the
anonymised documents retain the usefulness and utility of the original
documents. Put differently: researchers -- for whom the PSI itself is
often of no interest - can just as well work with the anonymised data
and still obtain the same results in statistical analyses and prediction
tasks. (2) For organisations and data-owners of sensitive data, it is
important that the process of anonymising data is successful and does
not itself introduce additional risks to the data. To mitigate any
concerns on that level, Textwash works fully offline and the data never
leave the user's system.

\hypertarget{core-findings}{%
\subsection{Core findings}\label{core-findings}}

This paper evaluated the text anonymisation software Textwash along
three criteria: a technical, an information loss, and a de-anonymisation
evaluation (see (Mozes and Kleinberg 2021)).

\hypertarget{technical-evaluation-2}{%
\subsubsection{Technical evaluation}\label{technical-evaluation-2}}

We assessed how many of the categories of phrase deemed meaningful to
anonymise text data were correctly identified by the underlying model on
unseen data. The weighted average of the F1 score over all categories
was high (0.93), suggesting that the model is able to identify the
categories correctly and makes the correct decision in the vast majority
of cases (weighted recall and precision: 0.93). There is some variation
across categories with ``occupation'' (\(F1 = 0.52\)) and
``other\_identifying\_attribute'' (\(F1 = 0.69\)) being at the lower end
of the per-category performances. Although this results in some
mismatching (e.g., an organisation identified as a location), it has no
adverse effect on the anonymisation since even misclassified categories
are removed. As a whole, the technical evaluation indicated that the
tool is comparable to state-of-the-art entity recognition
systems.\footnote{See
  \url{https://paperswithcode.com/sota/named-entity-recognition-ner-on-conll-2003}.}

\hypertarget{information-loss-2}{%
\subsubsection{Information loss}\label{information-loss-2}}

We used two common NLP tasks to test the information loss criterion,
which is further subdivided into utility loss and construct loss. First,
for utility loss, we compared the prediction performances for a popular
movie review dataset with those for the same but anonymised dataset. The
difference in performance can be attributed to the loss of information
due to the anonymisation procedure. Here, the data suggest negligible
loss (\textless{} 1\%). It is worth noting, however, that we focused on
sentiment analysis as a downstream task which may be less affected by
the removal of potentially sensitive information than other tasks.
Future work could examine various downstream tasks for a more
comprehensive picture. Second, we evaluated the construct loss by
testing for statistical differences in part-of-speech tag frequencies
between original and anonymised data. We used the person description
data collected for the motivated intruder testing and found statistical
evidence in favour of the hypothesis that the anonymisation procedure
does not introduce frequency differences in relevant POS tags. We do
observe considerable differences for POS tags that are directly removed
by the anonymisation model (e.g., pronouns). These differences vanish
when the replacements are mapped back to their corresponding POS tag
(e.g., {[}PRONOUN\_1{]} --\textgreater{} POS tag `PRON').

In summary, the information loss evaluation suggested that, for popular
research tasks (sentiment, POS tagging), the difference between raw and
anonymised data is negligible, so the conclusions that can be derived
from anonymised text data are identical to those from the raw, original
data.

\hypertarget{de-anonymisation-3}{%
\subsubsection{De-anonymisation}\label{de-anonymisation-3}}

De-anonymisation was introduced as the litmus test for the tool and
assessed with a motivated intruder procedure. We collected a set of
highly detailed person descriptions for individuals of varying levels of
fame. These descriptions were then presented in an anonymised form to
human intruders tasked with the identification of identities. The
results show that even the world's most famous individuals' descriptions
are rendered anonymous in 82\% (Study 1) and 74\% (Study 2) of the
cases. This finding is noteworthy because what would normally be
considered privileged information (e.g., where someone was born, their
partners) is in the public domain and often common knowledge for all
these very famous individuals. This is not the case for the typical
research context (e.g., interview transcripts, diary data) or envisioned
data sharing (e.g., police reports, health records), so the more
realistic benchmarks are the findings for semifamous and, even more so,
fictitious persons. For these, the descriptions are rendered anonymous
in 98\%-99\% of the cases.

We also explored how, in the cases where an intruder succeeded, they
were able to identify a very famous person. The findings here illuminate
that it is highly specific details that can be used only in combination
with public, privileged information for an educated guess. That kind of
information leakage shows one of the challenges of anonymising
qualitative data but does not threaten the anonymisation quality for the
typical use case.

\hypertarget{the-need-for-transparent-empirical-evaluation}{%
\subsection{The need for transparent, empirical
evaluation}\label{the-need-for-transparent-empirical-evaluation}}

To date, we are not aware of any text anonymisation tools evaluated to
the same rigour as we did in this paper. We encourage others who develop
anonymisation tools to follow suit and present more comprehensive
empirical validation; this paper can serve as a guideline for an
extensive procedure. At a minimum, a user of text anonymisation software
should insist on an adequate empirical validation of the software they
intend to use in terms of both anonymisation success and the
characteristics of the output text. The datasets produced in the studies
of the current paper are publicly available and can be used as a
benchmark for other anonymisation tools. Similarly, a mere technical
evaluation is insufficient: a tool could perform well on technical
performance metrics but fail with the essential task of protecting the
identity of individuals. The proposed motivated intruder testing as part
of the TILD criteria (Mozes and Kleinberg 2021) is a way to assess the
de-anonymisability of text data.

\hypertarget{implications}{%
\subsection{Implications}\label{implications}}

The availability of a validated text anonymisation tool has implications
on various levels.

\emph{Research:} With the rise of hybrid disciplines such as
computational social science, there is an increased need to break down
disciplinary silos. For text data, methodological advances typically
come from natural language processing and machine learning research
requiring large datasets (e.g., to train language models). Consequently,
many of the advancements in these areas are based on and applied to
easy-to-get large-scale data (e.g., Twitter data with all its known
problems (Morstatter et al. 2013; Pfeffer, Mayer, and Morstatter 2018)).
This may mean that research activity is biased towards topics that are
most amenable for study rather than those that are most meaningful or
important. With automated anonymisation, large datasets currently not in
use due to the hurdle of manual anonymisation can become public
datasets. Ideally, this would bring topical datasets (e.g., police
reports, health records) closer to the computational methods that have
advanced what we can learn from text data.

\emph{Data-sharing:} The dilemma between data sharing and the privacy
protection of research participants has long been considered an
either-or question, often decided in favour of not sharing data. With
the anonymisation tool introduced in this paper, that dilemma can be
resolved efficiently and effectively removing the conflict between these
two desirable open science practices (data sharing and privacy
protection). Researchers and organisations can now share data without
violating privacy or data protection guidelines. Likewise, science data
archives can now fulfil their mission of making data publicly available
-- particularly for studies funded by public research councils - without
risking privacy violations.

\emph{Open science:} The umbrella of all of the above-listed ways in
which automated, fast, and validated text anonymisation improves
procedures is open science. On the one hand, anonymisation tools enable
researchers who were, until now, understandably hesitant about sharing
qualitative text data to make their data public and thereby meet the
mandates of an increasing number of journals and research funding
agencies. It is also beneficial for the reliability of science because
by sharing more data, researchers can pool datasets and thereby increase
the sizes thereof, which can ameliorate the controversy of small sample
sizes (Yarkoni and Westfall 2017) and statistically under-powered
studies (Maxwell 2004; Card et al. 2020). Ultimately, having more data
available and large sample sizes will improve the reproducibility of
research.

\hypertarget{limitations-and-outlook}{%
\subsection{Limitations and outlook}\label{limitations-and-outlook}}

Despite the promise of our evaluation results, a few points merit
attention. First, even though the human intruder testing showed that
practically no individual that is not very famous could be identified
after detailed person descriptions are presented in anonymised form, the
question of `how good is good enough?' remains. We argue that in most
contexts, the performances obtained with the current Textwash tool are
sufficient to justify it as anonymisation, comparable to what a human
anonymiser would be able to do in considerably more time. Here, future
work could experimentally compare automated anonymisation approaches,
including Textwash, with manual human anonymisation. Improving the tool
is desirable regardless of the promising findings presented here. Future
work in this area could include an active learning module into the
Textwash pipeline: users are presented with a small number of samples of
documents after anonymisation and provide feedback on the quality of the
anonymised document. That way, the model can actively learn (and hence
optimise for) what desirable anonymisation is.

Second, the current procedure assumed that all documents are equally
anonymisable and that one degree of anonymisation fits all contexts. To
further improve the tool's performance, future work could seek to
identify an a priori document risk score. Using supervised machine
learning, one could train a higher-level model to estimate the degree to
which Textwash can anonymise data successfully, and then decide to
submit `difficult-to-anonymise' documents to human review. A related
future improvement could lie in adjusting the level of anonymisation
required: the stricter the anonymisation need, the lower one could set
the probability threshold of the underlying machine learning model to
remove a token. While the approach taken here resulted in little
information loss, the degree of removed tokens and the usefulness and
utility of anonymised data are inevitably a trade-off, and the user
could tailor this to their needs.

Third, one option not explored in the current paper is a
token-consistent replacement approach. Rather than replacing ``London''
with {[}LOCATION\_1{]}, we could replace it with ``Madrid.'' Such an
approach would likely improve the information loss criteria and possibly
the de-anonymisation rate. However, we desisted from that procedure as
it introduces additional complexity because it requires retaining
semantic relations between tokens. For example, an original sequence of
``{[}\ldots{]} lived in the capital of Spain, Madrid, and plays for the
local team Atletico Madrid'' would require a token-consistent
replacement where the relationship of country, capital city and local
football club is retained. That complexity could be solved with external
knowledge bases (Ji et al. 2022) in the future but is out of the scope
of this current work.

\hypertarget{conclusion}{%
\section{Conclusion}\label{conclusion}}

Text anonymisation is an important pillar of open science efforts and
has until now been insufficiently addressed for large-scale purposes.
This paper introduced the Textwash tool that allows researchers,
data-owners and individuals to automatically remove potentially
sensitive information from text data, thereby enabling them to share
data without compromising data privacy.

\hypertarget{acknowledgments}{%
\section{Acknowledgments}\label{acknowledgments}}

This work was supported by a Concept Grant from SAGE Ocean.

\hypertarget{full-author-affiliations}{%
\section{Full author affiliations}\label{full-author-affiliations}}

\begin{itemize}
\tightlist
\item
  Bennett Kleinberg, corresponding author: Department of Methodology \&
  Statistics, Tilburg University, The Netherlands; Department of
  Security and Crime Science, University College London, UK.Contact:
  \href{mailto:bennett.kleinberg@tilburguniversity.edu}{\nolinkurl{bennett.kleinberg@tilburguniversity.edu}}
\item
  Toby Davies: Department of Security and Crime Science, University
  College London, UK. Contact:
  \href{mailto:toby.davies@ucl.ac.uk}{\nolinkurl{toby.davies@ucl.ac.uk}}
\item
  Maximilian Mozes: Department of Computer Science \& Dawes Centre for
  Future Crime, University College London, UK. Contact:
  \href{mailto:maximilian.mozes@ucl.ac.uk}{\nolinkurl{maximilian.mozes@ucl.ac.uk}}
\end{itemize}

\hypertarget{references}{%
\subsection*{References}\label{references}}
\addcontentsline{toc}{subsection}{References}

\hypertarget{refs}{}
\begin{CSLReferences}{1}{0}
\leavevmode\hypertarget{ref-adams2019anonymate}{}%
Adams, Allison, Eric Aili, Daniel Aioanei, Rebecca Jonsson, Lina
Mickelsson, Dagmar Mikmekova, Fred Roberts, Javier Fernandez Valencia,
and Roger Wechsler. 2019. {``AnonyMate: A Toolkit for Anonymizing
Unstructured Chat Data.''} In \emph{Proceedings of the Workshop on NLP
and Pseudonymisation}, 1--7.

\leavevmode\hypertarget{ref-benoit_spacyr_2017}{}%
Benoit, Kenneth, and Akitaka Matsuo. 2017. {``Spacyr: {R} Wrapper to the
Spacy {NLP} Library.''} \url{https://CRAN.R-project.org/package=spacyr}.

\leavevmode\hypertarget{ref-berg-etal-2019-building}{}%
Berg, Hanna, Taridzo Chomutare, and Hercules Dalianis. 2019. {``Building
a de-Identification System for Real {S}wedish Clinical Text Using
Pseudonymised Clinical Text.''} In \emph{Proceedings of the Tenth
International Workshop on Health Text Mining and Information Analysis
(LOUHI 2019)}, 118--25. Hong Kong: Association for Computational
Linguistics. \url{https://doi.org/10.18653/v1/D19-6215}.

\leavevmode\hypertarget{ref-boyd_natural_2021}{}%
Boyd, Ryan L., and H. Andrew Schwartz. 2021. {``Natural {Language}
{Analysis} and the {Psychology} of {Verbal} {Behavior}: {The} {Past},
{Present}, and {Future} {States} of the {Field}.''} \emph{Journal of
Language and Social Psychology} 40 (1): 21--41.
\url{https://doi.org/10.1177/0261927X20967028}.

\leavevmode\hypertarget{ref-card2022computational}{}%
Card, Dallas, Serina Chang, Chris Becker, Julia Mendelsohn, Rob Voigt,
Leah Boustan, Ran Abramitzky, and Dan Jurafsky. 2022. {``Computational
Analysis of 140 Years of US Political Speeches Reveals More Positive but
Increasingly Polarized Framing of Immigration.''} \emph{Proceedings of
the National Academy of Sciences} 119 (31): e2120510119.

\leavevmode\hypertarget{ref-card2020little}{}%
Card, Dallas, Peter Henderson, Urvashi Khandelwal, Robin Jia, Kyle
Mahowald, and Dan Jurafsky. 2020. {``With Little Power Comes Great
Responsibility.''} \emph{arXiv Preprint arXiv:2010.06595}.

\leavevmode\hypertarget{ref-charlatanR}{}%
Chamberlain, Scott, and Kyle Voytovich. 2020. \emph{Charlatan: Make Fake
Data}. \url{https://CRAN.R-project.org/package=charlatan}.

\leavevmode\hypertarget{ref-bnc2007british}{}%
Consortium, BNC, and others. 2007. {``British National Corpus.''}
\emph{Oxford Text Archive Core Collection}.

\leavevmode\hypertarget{ref-devlin2018bert}{}%
Devlin, Jacob, Ming-Wei Chang, Kenton Lee, and Kristina Toutanova. 2018.
{``Bert: Pre-Training of Deep Bidirectional Transformers for Language
Understanding.''} \emph{arXiv Preprint arXiv:1810.04805}.

\leavevmode\hypertarget{ref-di2018towards}{}%
Di Cerbo, Francesco, and Slim Trabelsi. 2018. {``Towards Personal Data
Identification and Anonymization Using Machine Learning Techniques.''}
In \emph{European Conference on Advances in Databases and Information
Systems}, 118--26. Springer.

\leavevmode\hypertarget{ref-finkel2005incorporating}{}%
Finkel, Jenny Rose, Trond Grenager, and Christopher D Manning. 2005.
{``Incorporating Non-Local Information into Information Extraction
Systems by Gibbs Sampling.''} In \emph{Proceedings of the 43rd Annual
Meeting of the Association for Computational Linguistics (ACL'05)},
363--70.

\leavevmode\hypertarget{ref-francopoulo2007tagparser}{}%
Francopoulo, Gil. 2007. {``TagParser: Well on the Way to ISO-Tc37
Conformance.''}

\leavevmode\hypertarget{ref-francopoulo2020anonymization}{}%
Francopoulo, Gil, and Léon-Paul Schaub. 2020. {``Anonymization for the
GDPR in the Context of Citizen and Customer Relationship Management and
NLP.''} In \emph{Workshop on Legal and Ethical Issues (Legal2020)},
9--14. ELRA.

\leavevmode\hypertarget{ref-gentzkow2019text}{}%
Gentzkow, Matthew, Bryan Kelly, and Matt Taddy. 2019. {``Text as
Data.''} \emph{Journal of Economic Literature} 57 (3): 535--74.

\leavevmode\hypertarget{ref-hassan2019automatic}{}%
Hassan, Fadi, David Sánchez, Jordi Soria-Comas, and Josep
Domingo-Ferrer. 2019. {``Automatic Anonymization of Textual Documents:
Detecting Sensitive Information via Word Embeddings.''} In \emph{2019
18th IEEE International Conference on Trust, Security and Privacy in
Computing and Communications/13th IEEE International Conference on Big
Data Science and Engineering (TrustCom/BigDataSE)}, 358--65. IEEE.

\leavevmode\hypertarget{ref-information_commissioners_office_anonymisation_2012}{}%
Information Commissioner's Office. 2012. {``Anonymisation: Managing Data
Protection Risk Code of Practice.''}
\url{https://ico.org.uk/media/for-organisations/documents/1061/anonymisation-code.pdf}.

\leavevmode\hypertarget{ref-ji_survey_2022}{}%
Ji, Shaoxiong, Shirui Pan, Erik Cambria, Pekka Marttinen, and Philip S.
Yu. 2022. {``A {Survey} on {Knowledge} {Graphs}: {Representation},
{Acquisition}, and {Applications}.''} \emph{IEEE Transactions on Neural
Networks and Learning Systems} 33 (2): 494--514.
\url{https://doi.org/10.1109/TNNLS.2021.3070843}.

\leavevmode\hypertarget{ref-kleinberg2017web}{}%
Kleinberg, Bennett, and Maximilian Mozes. 2017. {``Web-Based Text
Anonymization with Node. Js: Introducing NETANOS (Named Entity-Based
Text Anonymization for Open Science).''} \emph{Journal of Open Source
Software} 2 (14): 293.

\leavevmode\hypertarget{ref-kleinberg2020measuring}{}%
Kleinberg, Bennett, Isabelle van der Vegt, and Maximilian Mozes. 2020.
{``Measuring Emotions in the COVID-19 Real World Worry Dataset.''} In
\emph{Proceedings of the 1st Workshop on NLP for COVID-19 at ACL 2020}.

\leavevmode\hypertarget{ref-lewis2020pretrained}{}%
Lewis, Patrick, Myle Ott, Jingfei Du, and Veselin Stoyanov. 2020.
{``Pretrained Language Models for Biomedical and Clinical Tasks:
Understanding and Extending the State-of-the-Art.''} In
\emph{Proceedings of the 3rd Clinical Natural Language Processing
Workshop}, 146--57.

\leavevmode\hypertarget{ref-liu2019roberta}{}%
Liu, Yinhan, Myle Ott, Naman Goyal, Jingfei Du, Mandar Joshi, Danqi
Chen, Omer Levy, Mike Lewis, Luke Zettlemoyer, and Veselin Stoyanov.
2019. {``Roberta: A Robustly Optimized Bert Pretraining Approach.''}
\emph{arXiv Preprint arXiv:1907.11692}.

\leavevmode\hypertarget{ref-maas2011learning}{}%
Maas, Andrew, Raymond E Daly, Peter T Pham, Dan Huang, Andrew Y Ng, and
Christopher Potts. 2011. {``Learning Word Vectors for Sentiment
Analysis.''} In \emph{Proceedings of the 49th Annual Meeting of the
Association for Computational Linguistics: Human Language Technologies},
142--50.

\leavevmode\hypertarget{ref-mamede2016automated}{}%
Mamede, Nuno, Jorge Baptista, and Francisco Dias. 2016. {``Automated
Anonymization of Text Documents.''} In \emph{2016 IEEE Congress on
Evolutionary Computation (CEC)}, 1287--94. IEEE.

\leavevmode\hypertarget{ref-maxwell2004persistence}{}%
Maxwell, Scott E. 2004. {``The Persistence of Underpowered Studies in
Psychological Research: Causes, Consequences, and Remedies.''}
\emph{Psychological Methods} 9 (2): 147.

\leavevmode\hypertarget{ref-morstatter_is_2013}{}%
Morstatter, Fred, Jürgen Pfeffer, Huan Liu, and Kathleen M. Carley.
2013. {``Is the {Sample} {Good} {Enough}? {Comparing} {Data} from
{Twitter}'s {Streaming} {API} with {Twitter}'s {Firehose}.''}
\emph{arXiv:1306.5204 {[}Physics{]}}, June.
\url{http://arxiv.org/abs/1306.5204}.

\leavevmode\hypertarget{ref-mozes2021no}{}%
Mozes, Maximilian, and Bennett Kleinberg. 2021. {``No Intruder, No
Validity: Evaluation Criteria for Privacy-Preserving Text
Anonymization.''} \emph{arXiv Preprint arXiv:2103.09263}.

\leavevmode\hypertarget{ref-mozes2021frequency}{}%
Mozes, Maximilian, Pontus Stenetorp, Bennett Kleinberg, and Lewis
Griffin. 2021. {``Frequency-Guided Word Substitutions for Detecting
Textual Adversarial Examples.''} In \emph{Proceedings of the 16th
Conference of the European Chapter of the Association for Computational
Linguistics: Main Volume}, 171--86.

\leavevmode\hypertarget{ref-pfeffer_tampering_2018}{}%
Pfeffer, Jürgen, Katja Mayer, and Fred Morstatter. 2018. {``Tampering
with {Twitter}'s {Sample} {API}.''} \emph{EPJ Data Science} 7 (1): 50.
\url{https://doi.org/10.1140/epjds/s13688-018-0178-0}.

\leavevmode\hypertarget{ref-radford2018improving}{}%
Radford, Alec, Karthik Narasimhan, Tim Salimans, and Ilya Sutskever.
2018. {``Improving Language Understanding by Generative Pre-Training.''}

\leavevmode\hypertarget{ref-romanov2019natural}{}%
Romanov, Aleksandr, Anna Kurtukova, Anastasia Fedotova, and Roman
Meshcheryakov. 2019. {``Natural Text Anonymization Using Universal
Transformer with a Self-Attention.''} In \emph{Proceedings of the III
International Conference on Language Engineering and Applied Linguistics
(PRLEAL-2019), Saint Petersburg, Russia}, 22--37.

\leavevmode\hypertarget{ref-rouder2009bayesian}{}%
Rouder, Jeffrey N, Paul L Speckman, Dongchu Sun, Richard D Morey, and
Geoffrey Iverson. 2009. {``Bayesian t Tests for Accepting and Rejecting
the Null Hypothesis.''} \emph{Psychonomic Bulletin \& Review} 16 (2):
225--37.

\leavevmode\hypertarget{ref-salganik_bit_2019}{}%
Salganik, Matthew J. 2019. \emph{Bit by Bit: {Social} Research in the
Digital Age}. Princeton, NJ: Princeton University Press.

\leavevmode\hypertarget{ref-socher2013recursive}{}%
Socher, Richard, Alex Perelygin, Jean Wu, Jason Chuang, Christopher D
Manning, Andrew Y Ng, and Christopher Potts. 2013. {``Recursive Deep
Models for Semantic Compositionality over a Sentiment Treebank.''} In
\emph{Proceedings of the 2013 Conference on Empirical Methods in Natural
Language Processing}, 1631--42.

\leavevmode\hypertarget{ref-NIPS2017_3f5ee243}{}%
Vaswani, Ashish, Noam Shazeer, Niki Parmar, Jakob Uszkoreit, Llion
Jones, Aidan N Gomez, Łukasz Kaiser, and Illia Polosukhin. 2017.
{``Attention Is All You Need.''} In \emph{Advances in Neural Information
Processing Systems}, edited by I. Guyon, U. Von Luxburg, S. Bengio, H.
Wallach, R. Fergus, S. Vishwanathan, and R. Garnett. Vol. 30. Curran
Associates, Inc.
\url{https://proceedings.neurips.cc/paper/2017/file/3f5ee243547dee91fbd053c1c4a845aa-Paper.pdf}.

\leavevmode\hypertarget{ref-wolf2019huggingface}{}%
Wolf, Thomas, Lysandre Debut, Victor Sanh, Julien Chaumond, Clement
Delangue, Anthony Moi, Pierric Cistac, et al. 2019. {``Huggingface's
Transformers: State-of-the-Art Natural Language Processing.''}
\emph{arXiv Preprint arXiv:1910.03771}.

\leavevmode\hypertarget{ref-xia-etal-2020-bert}{}%
Xia, Patrick, Shijie Wu, and Benjamin Van Durme. 2020. {``Which *{BERT}?
{A} Survey Organizing Contextualized Encoders.''} In \emph{Proceedings
of the 2020 Conference on Empirical Methods in Natural Language
Processing (EMNLP)}, 7516--33. Online: Association for Computational
Linguistics. \url{https://doi.org/10.18653/v1/2020.emnlp-main.608}.

\leavevmode\hypertarget{ref-yarkoni_choosing_2017}{}%
Yarkoni, Tal, and Jacob Westfall. 2017. {``Choosing {Prediction} {Over}
{Explanation} in {Psychology}: {Lessons} {From} {Machine} {Learning}.''}
\emph{Perspectives on Psychological Science} 12 (6): 1100--1122.
\url{https://doi.org/10.1177/1745691617693393}.

\end{CSLReferences}

\end{document}